\documentclass{article}

\usepackage[utf8]{inputenc} 
\usepackage[T1]{fontenc}    
\usepackage{hyperref}       
\usepackage{url}            
\usepackage{booktabs}       
\usepackage{amsfonts}       
\usepackage{nicefrac}       
\usepackage{microtype}      
\usepackage{lipsum}		
\usepackage{graphicx}
\usepackage{natbib}
\usepackage{doi}

\usepackage{arxiv}

\title{\textbf{Oracle Analysis of Representations for Deep Open Set Detection}}

\author{
Risheek Garrepalli \\
School of EECS,Oregon State University\\
garrepar@oregonstate.edu \\
} 

\usepackage{subfiles}

\usepackage{algorithm}
\usepackage{algorithmic}
\usepackage{graphicx}

\usepackage{booktabs}
\usepackage{arydshln}
\usepackage{lscape} 
\usepackage{float} 

\begin{document}
\maketitle


\graphicspath{ {images/} }

\begin{abstract}
    
Deep learning (DL) image classifiers are being applied to many problems where incorrect decisions could be potentially disastrous. A critical weakness of existing DL image classifiers is that they are trained under a closed-world assumption, namely, that the set of classes on which they have trained is complete and no additional classes will be encountered at run time. The problem of detecting a novel class at run time is known as the \textit{Open Category} problem or Open Set Detection. 

A promising approach for solving the open category problem is to apply deep anomaly detection methods to notice when a run-time image is an outlier relative to the training images. To succeed, deep anomaly detection methods must solve two problems: (i) they must map the input images into a latent representation that contains enough information to detect the outliers, and (ii) they must learn an anomaly scoring function that can extract this information from the latent representation to identify the anomalies.  

Research in deep anomaly detection methods has progressed slowly. One reason may be that  most papers simultaneously introduce new representation learning techniques and new anomaly scoring approaches. The goal of this work is to improve this methodology by providing ways of separately measuring the effectiveness of the representation learning and anomaly scoring. 

This work makes two methodological contributions. The first is to introduce the notion of \textit{oracle anomaly detection} for quantifying the information available in a learned latent representation. The second is to introduce \textit{oracle representation learning}, which produces a representation that is guaranteed to be sufficient for accurate anomaly detection. These two techniques help researchers to separate the quality of the learned representation from the performance of the anomaly scoring mechanism so that they can debug and improve their systems. The methods also provide an upper limit on how much open category detection can be improved through better anomaly scoring mechanisms. The combination of the two oracles gives an upper limit on the performance that any open category detection method could achieve.

This work introduces these two oracle techniques and demonstrates their utility by applying them to several leading open category detection methods. The results show that improvements are needed in both representation learning and anomaly scoring in order to achieve good open category detection performance on standard benchmark image classification tasks.
\end{abstract}




\section{Introduction}

Deep learning (DL) image classifiers are being applied to many problems, 
such as recognizing obstacles in autonomous driving and classifying X-ray images in 
health care, where incorrect decisions could be potentially disastrous. A critical weakness of 
existing DL image classifiers is that they are trained under a closed-world assumption, namely, 
that the set of classes (obstacles, diseases) on which they have trained is complete and no additional 
classes will be encountered at run time. The problem of detecting a novel class at run time is 
known as the \textit{Open Category} problem. 

A promising approach for solving the open category problem is to apply deep anomaly detection methods
 to notice when a run-time image is an outlier relative to the training images. To succeed, deep anomaly
  detection methods must solve two problems. First, they must map the input images into a latent 
  representation that contains enough information to detect the outliers. Second, they must learn
   an anomaly scoring function that can extract this information from the latent representation to identify the anomalies.  

Research in deep anomaly detection methods has progressed slowly. One reason may be that  
most papers simultaneously introduce new representation learning techniques and new anomaly 
scoring approaches. The goal of this work is to improve this methodology by providing 
ways of separately measuring the effectiveness of the representation learning and anomaly scoring. 

This work makes two methodological contributions. The first is to introduce the notion 
of \textit{oracle anomaly detection} for quantifying the information available in a 
learned latent representation. The oracle anomaly detector is trained via supervised learning 
to distinguish between images belonging to known classes and images belonging to novel classes 
using only the learned latent representation as input. If the oracle anomaly detector performs 
well, this proves that the latent representation is sufficient to support anomaly detection 
and therefore researchers should focus on improving the anomaly scoring mechanism. If the 
oracle anomaly detector performs poorly, this indicates that researchers need to improve 
the methods for learning the latent representation.

The second contribution is \textit{oracle representation learning} in which the representation 
learning mechanism is trained on images from the novel classes in addition to images from the 
known classes. If a good oracle can be trained, this guarantees that the resulting latent representation
 is sufficient to detect the novel classes. This can verify that the oracle anomaly detector is working 
 correctly. If a good oracle cannot be trained, this suggests that the neural network architecture needs
  to be improved or that the images lack information sufficient for the classification task. 

These two techniques also help researchers measure the extent to which existing methods can be improved. 
Oracle anomaly detection gives an upper limit on how much open category detection could be improved through
 better anomaly scoring mechanisms. The combination of the two oracles gives an upper limit on the performance
  that any open category detection method could achieve.

This work introduces these two oracle techniques and demonstrates their utility by applying 
them to several leading open category detection methods. The results show that improvements are 
needed in both representation learning and anomaly scoring in order to achieve good open category 
detection performance on standard benchmark image classification tasks. Based on these results, at the time of this we work we adopted a new hybrid approach to representation learning that gives modest improvements in open category detection.

\subsection{Notation and Problem Definition}
The open category problem arises within the context of standard supervised learning. Let $S$ be a collection of labeled examples 
$D=\{(x_1,y_1), \ldots, (x_N,y_N)\}$ where each $x_i$ is an image and each $y_i$ is a class label. 
In our experiments, an image is a $3 \times w \times h$ tensor with 3 color channels (R, G, B), width $w$, and height $h$.
 We will denote the space of all images by $\mathcal{X}$. Each class label is drawn from the 
 set $\mathcal{Y} = \{1, \ldots, K, K+1, \ldots, K+U\}$, where $K$ denotes the number of known classes 
 and $U$ denotes the number of unknown classes. The set $S$ is partitioned into three subsets corresponding 
 to the training data $D_r$, the validation data $D_v$, and the test data $D_t$. All of the examples in $D_r$ and $D_v$
  have labels drawn from $\{1, \ldots, K\}$, whereas the examples in $D_t$ have labels drawn from the entire label set. 

A closed-world classifier $f: \mathcal{X} \mapsto \Delta_{K}$ is a function that maps from
 the space of images to the probability simplex over the $K$ known classes, which we denote by $\Delta_{K}$.
  In this work, the classifier $f$ is a deep neural network, and we denote the weights of this network by $\theta$.
   An open category detector is a binary classifier $g: \mathcal{X} \mapsto [0,1]$ such that $g(x)$ is the (predicted) 
   probability that the input image $x$ belongs to one of the unknown classes $K+1, \ldots, K+U$ versus belonging to one of the known classes.  
A multi-class open category detector is a multi-class classifier $h: \mathcal{X} \mapsto \Delta_{K+1}$ is a function that maps to the probability simplex over the $K+1$ outputs. The the first $k$ outputs specify the probability that the input belongs to one of the $K$ known classes, and the $K+1$st output gives the probability that the input belongs to an unknown class.

The open category problem is to learn an accurate $g$ or $h$ using only the training and validation data $D_r$ and $D_v$.

In this work, the neural network $f$ will be divided into two sub-networks: the encoder $E$ and the classifier $C$. The 
encoder maps from the image space $\mathcal{X}$ to a latent space $\mathcal{Z}$, which we will usually assume is a
 $d$-dimensional real vector space $\Re^d$. The classifier then maps from $\mathcal{Z}$ to $\Delta_K$. 

Similarly, the open category detector $g$ will be divided into the same encoder $E$ and a scoring 
function $S$, where $S: \mathcal{Z} \mapsto [0,1]$. Hence, both the classifier $f$ and the open category 
detector $g$ share the same latent space $\mathcal{Z}$.

The problem of learning a good open category detector $g$ is therefore decomposed into the problem of learning a 
good latent representation $\mathcal{Z} = E(\mathcal{X})$ and a good scoring function $S$.

\begin{figure}[ht]
    \begin{center}
    \includegraphics[scale=0.8]{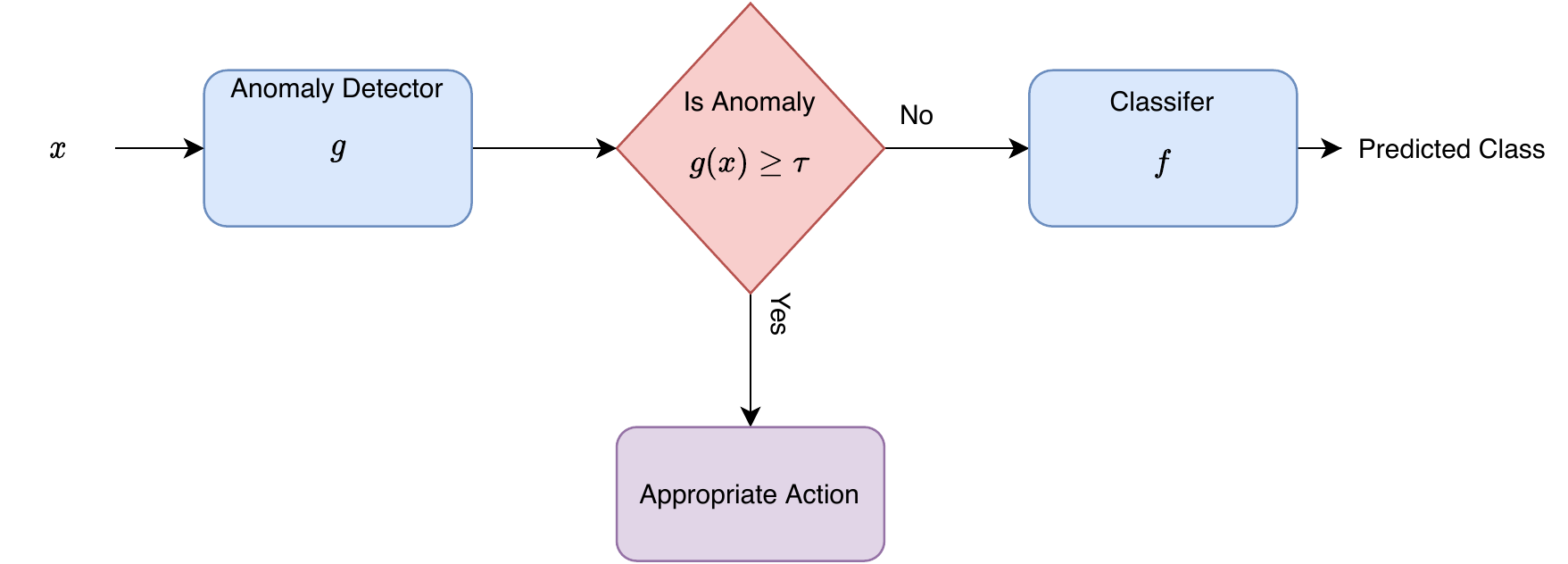}
    \end{center}
    \caption{Block diagram of Open Category Detection} 
    \label{overall_flowchart}
\end{figure}

 \indent The setup of open category detection and classification is illustrated in Figure~\ref{overall_flowchart}. Given a test query $x$, we first evaluate $g(x)$ and compare it to an anomaly score threshold $\tau$. If $g(x)\geq \tau$, $x$ is declared to belong to an unknown class, and an appropriate action (e.g., raising an alarm) is taken. Otherwise, the closed-world classifier $f$ is applied to predict a probability distribution over the known classes.

\subsection{Evaluation Metrics}
The performance of an open category detector $g$ is evaluated in the same way as a binary classifier.
 We employ the area under the ROC curve (AUC) \cite{roc-curve} as our primary metric, 
 as it provides a calibration-free measure of detection performance.



\subsection{Organization}

The remainder of this report is organized as follows.
Section 2 reviews previous work on representation learning and anomaly scoring for deep networks. In Section 3 we define different representations and anomaly measures considered in this report.
Section 4 introduces Oracle anomaly detection and Oracle representational learning. It then presents oracle experiments to understand what representations are effective for open set detection and what anomaly measures best capture the useful signal for open category detection.


\section{Related Problem settings} A variety of prior work has addressed variants of open category detection in low dimensional or featurized settings. This includes work on formalizing the concept of ``open space" to characterize regions of the feature space outside of the support of the training set \cite{6365193}. There is also a setting known as \textit{Open world recognition} (ODSN) \cite{DBLP:journals/corr/BendaleB15} that combines the problem of multi-class open set detection with the problem of incremental learning. Images predicted to belong to unknown classes are labeled via a labeling process (e.g., requesting a label from an expert) and then an incremental learning process updates the classifier to recognize the new category.

Out-of-distribution detection (OOD) is another closely related problem setting that is defined primarily by its methodology. As with open category detection, the training data consist of examples from $K$ known classes and the test data is a mix of data from known and unknown classes. However, the data from unknown classes is drawn from a different data set collected in a different way (e.g., under different imaging and image processing conditions). Recent work in this setting includes \cite{hendrycks17baseline} and \cite{liang2018enhancing}. The change in data collection conditions may make OOD problems easier than open category problems, but no systematic study comparing these settings has been conducted.

\cite{bulusu2020anomalous} provide an up to date survey on anomaly detection in deep learning.

\subsection{Discriminative Representations}
Deep learning has demonstrated excellent performance on image classification and other domains using supervised learning. The task is often formulated as a multi-class learning, and the models are trained to minimize the cross-entropy loss, that is to maximize the likelihood under a multinomial probability model (softmax). 
Architectures such as ResNet \cite{7780459} exhibit reliable performance in image domains. 

Another approach to supervised learning is metric learning. These train the latent representation so that images in the same class have nearby latent representations and images in different classes have latent representations that are well-separated. Three of the best approaches for metric learning are 
contrastive loss \cite{1640964}, triplet loss \cite{10.1007/978-3-319-24261-3_7}, and prototypical networks \cite{NIPS2017_6996}.


\subsubsection{Anomaly Measures with Discriminative Representations}
\indent \textbf{Low dimensional algorithms:} The problem of anomaly detection in low-dimensional spaces with meaningful features has been studied extensively. Many different algorithms have appeared in the literature and in a benchmarking study \cite{emmott2013},
the authors found the Isolation Forest \cite{liu2008} to be the most effective anomaly detector. Other competitive methods include 
LODA \cite{pevny:2015} and Local outlier factor (LOF) \cite{10.1145/342009.335388}. Based on this study, we will use Isolation Forest and LOF in this report. We can obtain a representation from any training methodology and apply these anomaly detection methods to detect anomalies.

\textbf{Sensitivity based methods:} ODIN \cite{DBLP:journals/corr/LiangLS17} proposes to increase the
difference between the maximum softmax scores of in-distribution and OOD samples by (i) calibrating the softmax scores by scaling the logits that feed into
softmax by a large constant (referred to as temperature) and (ii) pre-processing the input by perturbing it with the loss gradient. ODIN demonstrated that at
high temperature values, the softmax score for the predicted class is proportional to the relative difference between largest un-normalized output (logit) and the
remaining outputs (logits). Moreover, they showed empirically that the difference between the largest logit and the remaining logits is higher for the in-distribution
images than for the out-of-distribution images. 

\textbf{Confidence based Anomaly Measures}:  \cite{hendrycks17baseline} showed that the maximum output probability of a softmax-trained network (the confidence) provides a strong baseline anomaly detection method. Another closely-related anomaly measure is the entropy of the softmax probabilities. The entropy will be high if all of the output probabilities are similar, and it will be zero if one output class is predicted with probability 1.0. In a subsequent paper \cite{hendrycks2018deep}, the authors introduced a technique called outlier exposure. In addition to minimizing the softmax loss on the training data, outlier exposure attempts to maximize the softmax entropy on an auxiliary data set of known outliers. This gives good performance, but it is not applicable in the problems we are studying because such auxiliary data is not available.
In recent works such as \cite{vaze2022openset} they show that a good closed set classifier improves open set detection and this conclusion is consistent with this work, as in general openset detection struggles to distinguish between mis-classification within closed set samples vs open set instances. There has been further analysis by \cite{dietterich2022familiarity} where authors argue that max-logit based open set detection detects absence of familiarity rather than presence of novelty.


\indent \textbf{Disagreement in predictions:}
If we are able to capture uncertainty well, we would expect our predictive uncertainty to be high for novel queries coming from unknown classes. However, uncertainty estimation is non-trivial in deep learning.
One, expensive, way to estimate uncertainty is via ensembles. Deep Ensembles \cite{NIPS2017_7219} aggregate the predictions of 
individually trained networks to obtain final predictive uncertainty.
A related approach \cite{Yu_2019_ICCV} computes a  `classifier discrepancy'. Their  neural network has two predictive classifiers computed from a single shared latent representation. During training, they maximize the discrepancy of decision boundaries of the two classifiers while still requiring that they correctly classify the known classes. Like outlier exposure, they can train on auxiliary outlier data to ensure that the classifiers give very different output probabilities on these points. This is a simple example of a diversity-maximizing ensemble. 
 
Ensemble leave one out (ELOC) \cite{DBLP:journals/corr/abs-1809-03576} trains an ensemble of networks in which one known class is left out when training each network. The data for that class are used instead as an auxiliary data set for outlier exposure. The predicted probabilities of these networks are aggregated to obtain an uncertainty estimate for anomaly scoring. 

HyperGAN \cite{pmlr-v97-ratzlaff19a} is technique in which a generator network produces as output the weights of a classifier network. By invoking the HyperGAN many times, a large ensemble can be cheaply constructed and then evaluated to assess disagreement and obtain an anomaly score. 

Bayesian learning seeks to represent uncertainty explicitly as a posterior distribution over networks. Doing this exactly is intractable. An additional challenge is that it is not obvious how to place a prior distribution over networks because they are overparameterized and nonlinear. 

The most practical approximate Bayesian method is Monte Carlo dropout (MC-Dropout) \cite{JMLR:v15:srivastava14a}, which was originally introduced as a regularization method. In MC-Dropout, subsets of weights in a network are randomly set to zero (``dropped out'') during learning each minibatch to prevent overfitting. 
\cite{pmlr-v48-gal16} show that this same mechanism can be used during prediction to generate multiple predicted probability vectors. These can be interpreted as (approximate) samples from a posterior distribution over networks. In this work, we do not investigate Bayesian methods. 

\subsection{Generative Models} Representations learned by discriminative training should not perform as well as representations learned through generative models or auto-encoders. This hypothesis is based on the following analysis. 
Discriminative training seeks to extract the minimum information needed to determine the class of the object in the image. If there are $k$ classes, one needs only $\lceil \log k\rceil$ bits of information, so we expect that a classifier works by removing irrelevant information until only the information sufficient to perform the classification task remains. Mathematically, discriminative training learns $p(y|x)$. However, the task of open category detection would appear to require modeling $p(x)$ or $p(x|y)$, the density over the input space, in order to detect new points $x$ that fall outside the distribution of the known classes. 

\indent Generative models learn an implicit representation of $p(x|y)$ in the sense that we can sample images $x \sim p(x|y)$. Auto-encoders learn a latent representation $z = E(x)$ such that a generator $G$ can reconstruct $x$: $\hat{x} = G(z)$. If the reconstruction is accurate, then $z$ must not be losing any information about $x$. This is consistent with the manifold assumption which states that the true dimensionality of the space of images is much lower than its apparent (``ambient") dimensionality (the number of pixels $\times$ 3, for RGB images).

\subsubsection{Anomaly Measures with Generative Models}
By training an auto-encoder or a generative model, we hope that we are learning this manifold. Hence, we can then learn a probability distribution $p(z)$ and detect novel categories as outliers in the manifold. It is possible that the images $x$ from new categories do not lie in the learned manifold. When this occurs, $E(x)=z$ projects those images into the manifold $z$. But when the image is reconstructed, there will be a large difference between $x$ and $\hat{x} = G(z)$. This reconstruction error can also provide a signal for open category detection that exploits the failure of the network to model the anomalies.   

\indent Dominant models of autoencoders include  variational autoencoders (VAEs) \cite{DBLP:journals/corr/KingmaW13}, which  motivate formulation of autoencoders from a probabilistic perspective, Wasserstein Autoencoders (WAEs) \cite{tolstikhin2018wasserstein}, which use the Wasserstein distance instead of KL-divergence, and  regularized autoencoders (RAEs) \cite{Ghosh2020From}.  RAEs demonstrate that just regularizing the latent
space in deterministic autoencoders maintains much of good behavior of VAEs/WAEs while being significantly easier to train. We apply RAEs in this work. 

Deep density estimation methods provide another approach to anomaly detection. We performed some experiments with Masked Autoregressive Flow \cite{NIPS2017_6828} but did not find that it performed very well as an anomaly detector. A possible reason is that such deep generative models do not perform any dimensionality reduction and hence do not take advantage of the manifold structure of the data.
An advantage of deep density estimation methods is that they allow us to query $-\log P(x)$ whereas autoencoders require us to compute both $-\log P(z)$ and the reconstruction error.

There is work exploring generative models for open set detection such as C2AE \cite{Oza_2019_CVPR}, where the encoder of an autoencoder is a pretrained classifier and the decoder is trained discriminatively to maximize the reconstruction error for unknowns and minimize reconstruction error for knowns. During evaluation, reconstruction error is used as anomaly score to detect unknowns. Another interesting approach is LCVAE \cite{sun2020conditional}, which trains a Ladder VAE \cite{NIPS2016_6275} architecture instead of a vanilla autoencoder. The LCVAE encoder is jointly trained to be a classifier, and the latent representation is trained to perform well as class conditional Gaussian classifier. 

Another generative model approach is Open Set Learning with Counterfactual Images \cite{Neal_2018_ECCV} where GANs are used to generate synthetic examples, and a classifier is trained to reject these synthetic examples. During test time, the rejection probability score is employed to detect unknown.


\section{Representations and Anomaly Measures}

In this section we discuss various representations and anomaly measures considered in this work.

To understand the properties and biases of representations in the context of open set detection, we consider representations obtained under different objectives and training schemes in both supervised and unsupervised settings.

\subsection{Discriminative training}

\begin{figure}[ht]
    \begin{center}
    \includegraphics[scale=0.6]{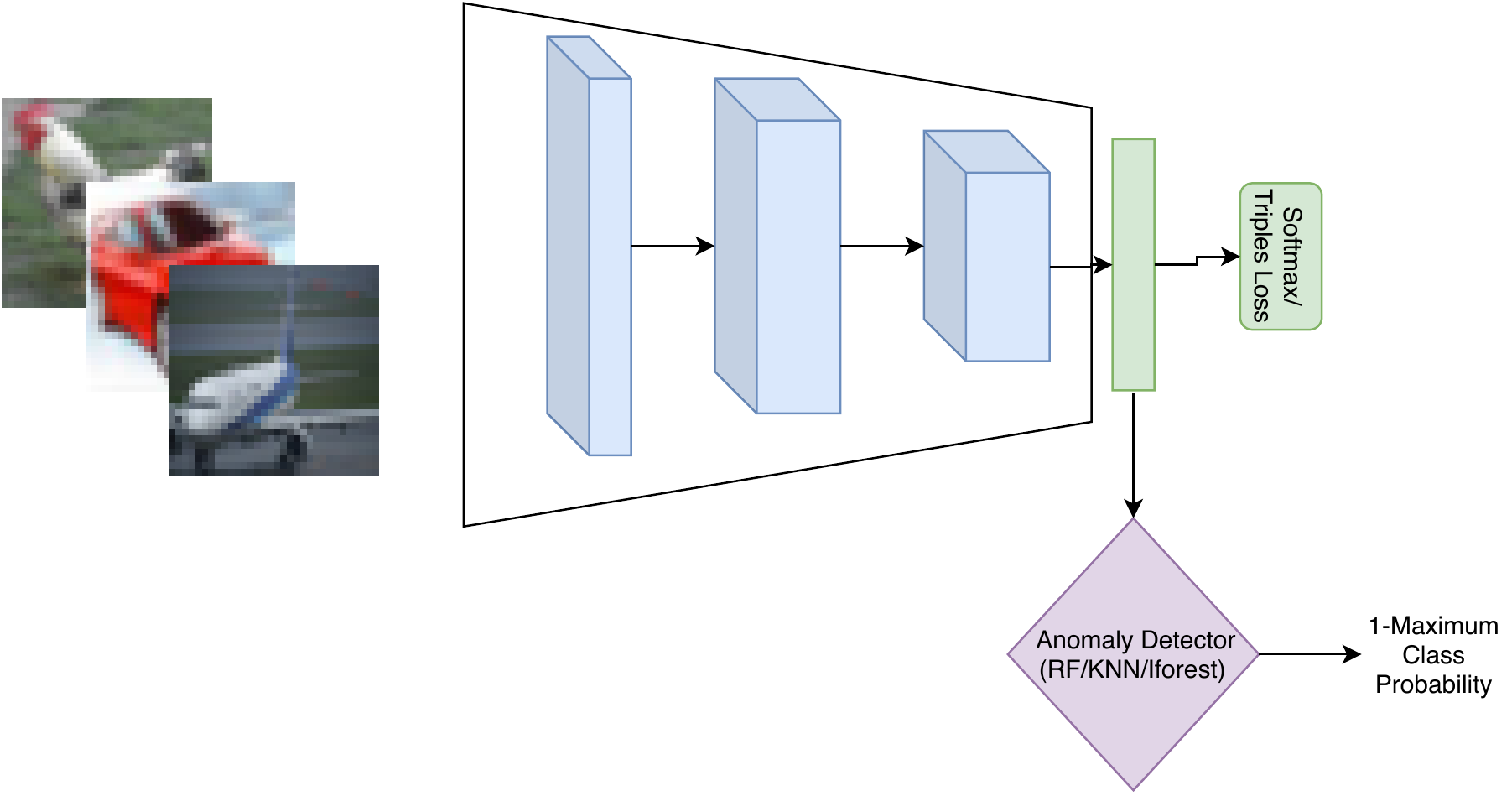}
    \end{center}
    \caption{Representations from Discriminative training} 
    \label{Discriminative-representations}
\end{figure}

Figure~\ref{Discriminative-representations} illustrates the basic setup for using discriminative training to learn a latent representation that can then be analyzed to score anomalies. The ``penultimate layer" of the network provides the representation. 

\subsubsection{Softmax}
As discussed in the introduction, let $K$ be number of known classes and $j$ be the class index. For any given input image $x$ we compute the representation $E(x)$ and then apply it through classifier to compute the logits and predictive probabilities $P(y|x)$.

Let $z_j$ be the final activation per class. The predicted conditional probabilities are computed via the softmax transformation:

\begin{equation}
    P_{sftmx}(y = j|x) = \frac{e^{z_j}}{\sum^{K}_{j=1}e^{z_j}}.
\end{equation}

The log-likelihood is also known as the cross-entropy, and it is minimized during DL training:

\begin{equation}
    L_{CE}(f(x;w),y) = - \sum^{K}_{j = 1} y \cdot \log p(y_{i}|x) + (1-y) \cdot \log (1-p(y_{i}|x)).
\end{equation}
    
\subsubsection{One-vs-Rest (Binary Cross-Entropy)}
Another way to train a multi-class classifier is using the one-vs-rest training methodology discussed above. To get predictions, we apply the logistic function to each class's corresponding activation.

\begin{equation}
    P(y_{j} = 1|x) = \frac{1}{1+e^{-z_j}}.
\end{equation}

The loss function sums the log-likelihood of each of the outputs. This is called the binary cross-entropy (BCE):
\begin{equation}
    L_{BCE}(f(x;w),y) = - \sum^{K}_{j = 1} y \cdot \log p(y_{j} = 1 |x) + (1-y) \cdot \log(1-p(y_{j} = 1|x)).
\end{equation}

\subsubsection{Metric Learning: Triplet Loss}
In deep metric learning, the input $x$ is passed through the neural network encoder $E$ to obtain an embedding, $z=E(x)$. A metric learning objective, such as the triplet loss (Equation \ref{Triplet}), is applied in the embedding space. The triplet loss is applied triplets consisting of three training examples: $(z_A,z_P,z_N)$, where $z_A$ is the ``anchor'', $z_P$ is an example that belongs to the same class as $z_A$ (``positive''), and $z_N$ is an example that belongs to a different class (``negative''). The loss is zero only if the distance $d(z_A,z_P)$ from the anchor to the positive is less than the distance $d(z_A,z_N)$ from the anchor to the negative by an amount of at least $\alpha$. It is not feasible or desirable to train on all possible triplets. Instead, a typical approach is to form one triplet for each training example $z_A$ in a mini-batch.
In practice instead of calculating loss on all possible triplets within a mini-batch usually top-k most difficult triplets are chosen to compute the loss, 
a practice known as ``hard negative mining''. In addition to the Triplet loss of Equation (\ref{Triplet}), the learned $z$ values are regularized to have unit norm, which stabilizes the optimization.

\begin{equation}\label{Triplet}
    L(z_A,z_P,z_N) = \max\{(d(z_A,z_P)-d(z_A,z_N) + \alpha),0\}.
\end{equation}

\subsection{Generative Representations}
\begin{figure}[ht]
    \begin{center}
    \includegraphics[scale=0.6]{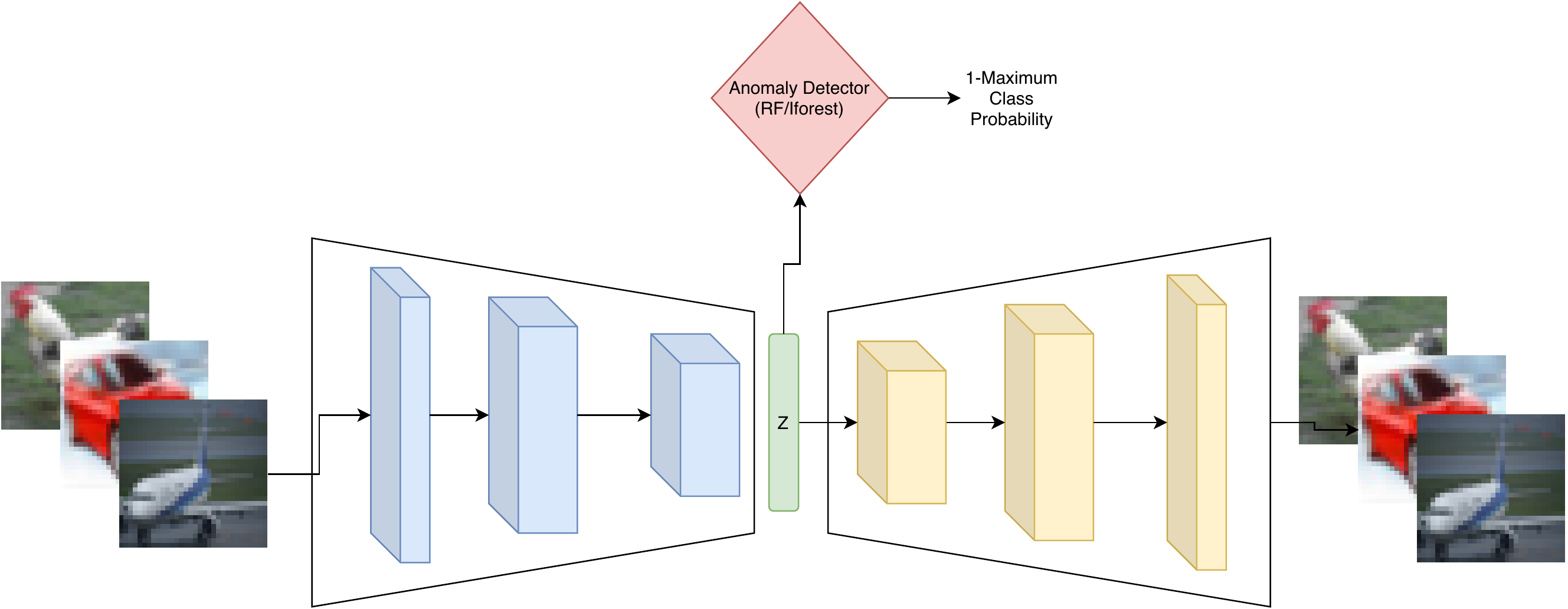}
    \end{center}
    \caption{Representations from Autoencoders} 
    \label{Generative-representation}
\end{figure}

Recall that an autoencoder consists of an encoder $z=E(x)$ and a decoder (or generator) $\hat{x}=G(z)$. The latent space $z$ must have relatively low dimension to force the encoder to learn a meaningful encoding of the image. For this reason, it is often referred to as the bottleneck layer, and it is usually fully connected to the adjacent layers. A typical setup for autoencoders is illustrated in figure \ref{Generative-representation}. The learning objective for deterministic autoencoder is defined as 
\begin{equation}
	L_{mse} = - \sum {\|x - \hat{x}\|^2}.
\end{equation}

We will also employ the regularized deterministic autoencoder (RAE) from \cite{Ghosh2020From}. This adds a squared norm penalty controlled by a regularization parameter $\lambda$:
\begin{equation}
    L_{RAE} = L_{mse} + \lambda \|z\|_{2}^2 
\end{equation}

\subsection{Hybrid Representations}

In addition to discriminative and generative representations, we also consider hybrid representations where the latent representation is shared between classifier and decoder. The training objective is a combination of the cross-entropy loss and reconstruction loss, and no regularization is applied to the shared representation.

\begin{equation}
    L_{hybrid}(f(x;w),y) = - \left\{\lambda  \sum \|x - \hat{x}\|^2 + \sum^{K}_{j = 1} \left[ y \cdot \log p(y_{i}|x) + (1-y) \cdot \log (1-p(y_{i}|x))\right] \right\}.
\end{equation}

\subsection{Different representations}
For any given input query $x$, we denote representation by $Z$ with a suffix indicating the training method.
\begin{itemize}
    \item The representation $Z_{sftmx}$ is the penultimate layer of a deep network obtained from softmax-based discriminative learning.
    \item The representation $Z_{ae}$ is the bottleneck layer of a deep autoencoder trained by using reconstruction loss.
    \item The representation $Z_{sftmx+ae}$ is the hybrid representation obtained by jointly training a shared representation to act as the penultimate layer of classifier and as the input to the decoder of an autoencoder.
    \item The representation $Z_{bce}$ is the penultimate layer of a deep network obtained via one-vs-rest discriminative learning minimizing the class binary cross entropy (BCE).
    \item The representation $Z_{bce\_finetuned}$ is the penultimate layer of a deep network obtained by fine tuning $Z_{sftmx}$ with BCE (one-vs-rest). That is, first $Z_{sftmx}$ is trained and then some additional training is performed to minimize the BCE objective.
    \item The representation $Z_{triplet}$ is the embedding layer of a network trained to minimize the triplet loss.
    \item The representation $Z_{triplet-1}$ is obtained using the triplet loss and then extracting the activations of 
    the layer of the network immediately prior to the final embedding layer.

\end{itemize}

\section{Anomaly Measures}

To detect open set inputs, we need a measure of how different a given query is compared to the training data. For each representation, we consider different anomaly signals that quantify the degree to which an point $x$ is anomalous.
In this section we describe the different anomaly signals that can be applied to each representation. In the following chapter, we will present experiments that compare these signals to determine which ones are most effective for anomaly detection. For each representation, we apply each the following anomaly measures when possible.

\subsection{Predictive entropy of classifier}
One of the signals we can use for anomaly detection is entropy of predictive probability distribution within the known classes. Given an input image $x$, let $z=E(x)$ be one of the latent representations $z_{sftmx}$, $z_{bce}$, or $z_{bce\_finetuned}$. Let $p(\hat{y}=k|z)$ be the predicted probability that $x$ belongs to class $k$. Then the entropy measures the amount of uncertainty in this distribution. It is computed as 

\begin{equation}
    H(y|x) = - \sum_{k=1}^K {p(\hat{y}=k|E(x)) \cdot \log p(\hat{y}=k|E(x))}
\end{equation}

\subsection{Maximum class probability}
Another anomaly signal we can consider is the negative of the confidence of the classifier. This is larger when the classifier is less confident.  It is defined as the negative of the maximum class probability:
\[
MaxP(y|x) = - \arg_k \max P(y=k|E(x)).
\]
We will add a subscript depending on the latent representation: $MaxP_{sftmx}$, $MaxP_{bce}$, and $MaxP_{bce\_finetuned}$.

\subsection{Maximum class logit}

A similar anomaly signal is the negative of the maximum of the logits, which are the unnormalized inputs to the softmax or the binary cross-entropy. Some authors argue that the unnormalized values contain more information about $x$ than the normalized probabilities. Let $\theta(k)$ be the weight vector of the logit for class $k$. That is, $\theta(k)^\top z$ computes this logit value. Then this anomaly signal is computed as 

\begin{equation}
     MaxC_{\theta}(x) = - \arg_k \max \theta(k)^\top E(x).
\end{equation}

\subsection{Autoencoder reconstruction error}
For a given image query $x$, the reconstruction error is computed as

\begin{equation}
    a_{mse}(x) = \frac{1}{h\cdot w}\sum_{u=1}^h \sum_{v=1}^w \|x[u,v] - \hat{x}[u,v]\|^2,
\end{equation}
where $a_{mse}$ denotes the squared error of the reconstructed image of width $w$ and height $h$ and $u$ indexes the rows and $v$ indexes the columns of the image.

\subsection{Anomaly detection Algorithms}
In addition to anomaly signals from classifiers and reconstruction error, we also train anomaly detectors
explicitly on $D_v$ on a given representation and use the anomaly score from trained anomaly 
detectors. 

\textbf{Isolation Forest:}
The Isolation Forest algorithm \cite{liu2008} ranks data points as anomalous if they are easily isolated (separated by the training data) by random axis parallel splits. 
An isolation forest is a set of isolation trees, where each isolation tree is
an extremely randomized decision tree. During training, each tree is grown according to the usual top-down splitting approach. Each node is determined by choosing a feature uniformly at random and then choosing a splitting threshold uniformly at random between the minimum and maximum observed values of that feature. The tree is grown until each training example is isolated in its own leaf. The isolation depth of a training example is equal to the number of internal nodes that are traversed between the root and the leaf. Similarly, a query point is said to be isolated at depth $h_t$ in tree $t$ if $h_t$ decision nodes are traversed between the root and the leaf node.  The \textit{anomaly score} $a_{iforest}(x)$ for data point $x$ in forest $F$ consisting of $T$ trees $t_1, t_2, . . . , t_T$ built from $n$ data points is defined as
\begin{equation}
    a_{iforest}(x) = 2^{-\frac{\frac{1}{T} \sum_{t=1}^T h_{t}(x)}{c(n)}},
\end{equation}
where $c(n)$ is the expected depth of a leaf in a random tree having $n$ leaves.

The intuition is that outliers are points that are easily isolated at random from the remaining data, so points with small average isolation depth are likely to be anomalies and so are assigned high anomaly scores. 

\textbf{Local Outlier Factor (LOF)} \cite{10.1145/342009.335388} computes the outlier score $a_{lof}$ of a point $x_i$ by computing its average distance to its `k' nearest neighbors. It normalizes this distance by computing the average distance of each of those neighbors to their `k' nearest neighbors. The intuition is that a point is believed to be more anomalous if it is significantly farther from its neighbors than they are from each other.
\section{Oracle Anomaly Detection}

In this section, we introduce our two methodological contributions: oracle anomaly detection and oracle representation learning.

\subsection{Oracle representation learning}
To estimate an empirical upper bound on anomaly detection for a given architecture and data set, we introduce oracle representational learning. To obtain an oracle latent representation, we train a network $f$ on all of the data $D_r \cup D_v \cup D_t$, that is on all the classes including the unknown classes in the test data. If a good oracle cannot be trained, this suggests that the neural network architecture needs to be improved or that the images lack information sufficient for the classification task. If a good oracle can be trained, this guarantees that the resulting latent representation is sufficient to detect the novel classes. This can verify that the oracle anomaly detector is working correctly.

The gap in performance between anomaly detection using an oracle representation and anomaly detection using a nominal representation helps us understand  various dependencies such as effect of a particular class being present in the training phase or the impact of the total number of training classes. 
These can also be useful for evaluating anomaly detection in few shot learning or active learning scenarios.

\subsection{Oracle anomaly detection}
To quantify the information available for anomaly detection in a learned latent representation, we introduce the concept of an oracle anomaly detector (``Oracle AD''). An oracle anomaly detector is a classifier trained via supervised learning to distinguish queries from known and unknown classes using only the latent representation as input. If the oracle anomaly detector performs well, this proves that the latent representation is sufficient to support anomaly detection and therefore research should focus on improving the anomaly measure. 

In this work we consider two kinds of oracle ADs: a binary classifier and a multi-class classifier. In later sections, we evaluate each representation with both binary and multi-class oracle ADs. A binary oracle anomaly detector $g$ is trained on the latent representations of the examples in the training and validation sets $D_r$ and $D_v$. Examples belonging to known classes are labeled as $0$ and examples belonging to unknown classes are labeled as $1$. 
Similarly, a multi-class oracle anomaly detector $h(z)$ is trained on the latent representations of the examples in $D_r$ and $D_v$. Examples from the known classes are labeled as $\{1, \ldots, K\}$, and examples from unknown classes are labeled as class 0.

\section{Experiments}

\subsection{Research questions}

\begin{itemize}
    \item RQ1: Which methods of representation learning give the best representations (as measured by oracle AD)?
    \item RQ2: How big is the gap between oracle AD and existing anomaly scoring methods?
    \item RQ3: How big is the gap between oracle representations (measured with oracle AD) and existing representation learning methods (also measured with oracle AD)?
\end{itemize}

We describe two sets of experiments performed to answer these research questions.
First we apply oracle anomaly detection to evaluate different representations for detecting anomalies. Then we apply Oracle representation learning and oracle AD to estimate the ``head room'' for anomaly detection.

\subsubsection{Datasets}
In this section, dataset construction for our open category detection experiment is described. We should note that the construction here is different from the application setting in which the training and validation data only contain examples from known classes. For oracle training purposes, we need examples of unknown classes in both the validation and test datasets. 

\begin{figure}[H]
    \begin{center}
    \includegraphics[scale=0.75]{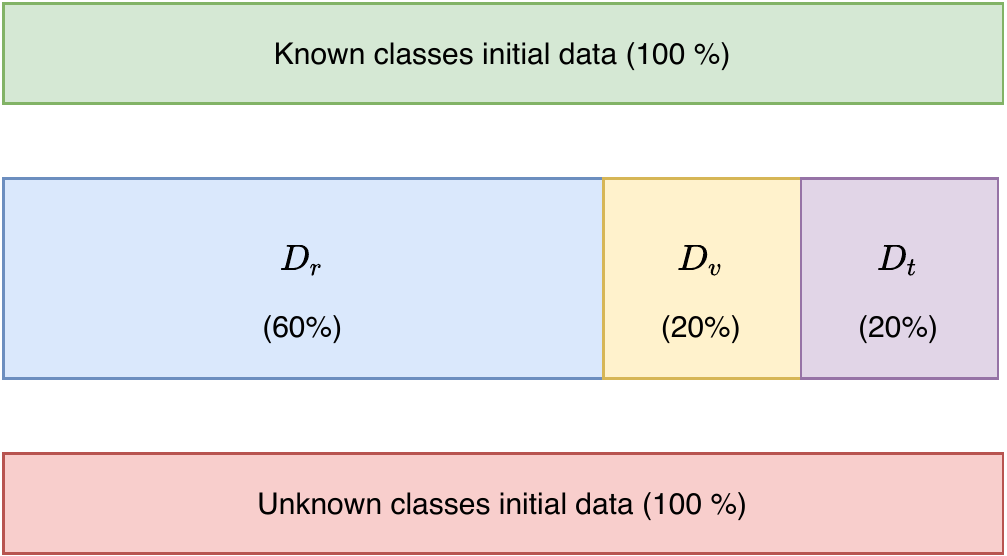}
    \end{center}
    \caption{Dataset construction and partition} 
    \label{Dataset split-openset}
\end{figure}

To construct our datasets, we first define which classes will be considered 
First, classes within each dataset are partitioned into disjoint sets of known and unknown classes. All of the examples belonging to known classes will be referred to as the ``known dataset'', and all examples belonging to unknown classes will be referred to as the ``unknown dataset''. 
For Oracle representational learning, we use samples from known and unknown classes of the validation data $D_v$ to train a binary classifier $g$ to distinguish known and unknown classes as illustrated in Figure \ref{Dataset split-openset} and described below.

\begin{itemize}
    \item A partition $D_r$ constitutes 60\% of known dataset to train the encoder $E$ and classifier $C$.
    \item A partition $D_v$ constitutes 20\% known dataset and an equal number of 
    samples drawn from the unknown dataset.
    \item A partition $D_t$ constitutes 20\% of the known dataset and an equal number of 
    samples drawn from the unknown dataset.
\end{itemize}

We evaluate open set classification performance on the MNIST, CIFAR10 and CIFAR100 datasets. The MNIST digit data set consists of ten classes of hand-written digits, and each instance is a 28x28 monochrome image. 
The CIFAR10 and CIFAR100 datasets consist of 10 and 100 classes of real-world objects. Each image is a 3-channel RBG image: 3*32x32. 
For MNIST and CIFAR10, we select 6 classes to be known and 4 classes to be unknown. For CIFAR100, we select 80 classes to be known and 20 classes to be unknown.

To measure performance, we compute the mean AUROC (with 95\% confidence intervals) from five different draws of known and unknown classes. When training neural networks, the seed is fixed to avoid the variance due to initialization. In case of training  oracle representations we only have one set of classes. Hence, we compute a confidence interval of AUROC by bootstrapping the final predictive probability scores of test set queries.
In the autoencoder experiments, we only conduct experiments on one known-unknown split and report AUROC without confidence intervals.

\subsubsection{Network architectures}

\begin{itemize}
    \item ResNet34 \cite{7780459} is used for all the experiments on the CIFAR10 and CIFAR100 datasets. For the autoencoder experiments, we again use ResNet34 but treat the global pool layer (512 features) as the latent representation layer (bottleneck) and pass it to the decoder. 
    \item For the CIFAR data sets, the architecture of the decoder begins with a linear layer with output size of $512\times 8\times 8$ followed by $3$ residual deconvolution blocks. Each block has $3$ deconvolutional layers with each layer consisting of 512 filters and a kernel size of 3. The reconstructed image $\hat{x}$ is computed by a final convolutional layer with a kernel size of 2.  In our experiments, we found that performance (of the learned representation and the reconstruction error) was improved by increasing the depth of the decoder and including skip connections. We applied spectral normalization to regularize each deconvolutional layer. All layers employ ReLU activation functions.
    \item For MNIST autoencoder experiments, we employed the architecture from  \cite{Ghosh2020From} as described in table \ref{mnist_nw}.
\end{itemize}

\begin{table}[ht]
    \begin{center}
    \caption{MNIST Network Architecture}
    \label{mnist_nw}
    \vskip 0.15in
    \begin{tabular}{l|rll}
    \hline
    \multicolumn{4}{c}{\textbf{Encoder or Classifier}}                                                                  \\
    \textbf{Layer} & \multicolumn{1}{l}{\textbf{Filter size}} & \textbf{Normalization} & \textbf{Activation} \\
    \cdashline{1-4} \\
    Conv1          & 128                                      & Batchnorm              & RelU                \\
    Conv2          & 256                                      & Batchnorm              & RelU                \\
    Conv3          & 512                                      & Batchnorm              & RelU                \\
    Conv4          & 1024                                     & Batchnorm              & RelU                \\
    Flatten        & \multicolumn{1}{l}{}                     &                        &                     \\
    Linear         & 64                                       &                        &                     \\
    \cdashline{1-4} 
    \multicolumn{4}{c}{\textbf{Decoder}}                                                                             \\
    \cdashline{1-4} \\
    \textbf{Layer} & \textbf{Filter size} & \textbf{Normalization} & \textbf{Activation} \\
    Linear         & 8*8*1024             & Batchnorm              & RelU                \\
    ConvT1         & 512                                      & Batchnorm              & RelU                \\
    ConvT2         & 256                                      & Batchnorm              & RelU                \\
    ConvT3         & 1                                        &                        &                    \\
    \hline
    \end{tabular}
    \end{center}
\end{table}

\subsubsection{Training and hyperparameters}

To estimate anomaly scores based on isolation Forest (iforest), we train a separate iforest for each class and then we combine all the individual isolation forest anomaly scores by taking a minimum of anomaly scores. To obtain similar performance when training on all classes together, we would need a very large isolation forest, which was not practical. To obtain a baseline anomaly score, we trained Local Outlier Factor (LOF) on the combined data from all classes.

For the oracle anomaly detector, we employ a random forest classifier with 1000 trees and compute the predictive probabilities of random forest to obtain oracle AUROC.

We encountered difficulty training a classifier for CIFAR100 with the BCE loss. We resolved this by initializing the network weights with values obtained by training CIFAR100 with the Softmax loss. We then trained with the BCE loss. This gives us the $Z_{bce\_finetuned}$ representation. 

To train all representations except triplet loss, we follow a training protocol where we train ResNet34 model for 200 epochs with mini-batches of size 128. We employ SGD with the Nesterov acceleration as our optimizer  
with the following learning rate (lr) schedule: The learning rate is set to $0.1$ for first 60 epochs, then $0.02$ for next 60 epochs, then $0.004$ for next 40 epochs, and finally to $0.0008$ for the last 40 epochs. For training with the hybrid loss $L_{hybrid}$ in Equation 3.8, we use $\lambda = 0.9$ for CIFAR10 and $\lambda = 0.005$ for CIFAR100.

For training with the triplet loss, we consider mini-batches of size 150, and we choose the 3000 most difficult triplets (via top-K mining) to compute the loss. We train the ResNet34 model for 225 epochs and use the checkpoint that gave the lowest triplet loss on the validation set. We employ the Adam optimizer with a learning rate of $3\times 10^{-4}$.

We found that both auto-encoders and metric learning methods were difficult to tune. Consequently, we report auto-encoder results for only the MNIST and CIFAR10 datasets and metric learning for only CIFAR10.

\section{Results}

Tables~\ref{oracle-cifar10} and \ref{oracle-cifar100} present the main results of the experiments of CIFAR100 and CIFAR10. The left column ``AD Treatment'' contrasts three settings. The nominal representation (``Nominal Rep'') treatment corresponds to the standard configuration in which the network is trained using one of loss functions described in the previous chapter. The ``Oracle AD (Nominal Rep)'' treatment trains an oracle anomaly detector on the nominal representation. Finally, the ``Oracle AD (Oracle Rep)'' first trains the oracle representation (treating all classes as known) and then trains an oracle anomaly detector on this representation. Hence, the performance numbers for the Nominal Rep are the AUROC values achievable in practice. The Oracle AD (Nominal Rep) treatment estimates the best performance that can be achieved with an ideal anomaly measure. And the Oracle AD (Oracle Rep) gives the best performance that could be achieved with this network architecture given an ideal representation and ideal anomaly measure. 

\begin{table}[ht]
    \centering
    \caption{Nominal Representations and Oracle AD on CIFAR100}
    \label{oracle-cifar100}
    \vskip 0.15in
    \begin{tabular}{cl|lll}
    \hline
        
        \multicolumn{1}{l}{}             & Training loss & Softmax         & BCE\_finetuned                                                       & Hybrid            \\
        \multicolumn{1}{l}{AD Treatment} &                 & $Z_{sftmx}$  & $Z_{bce\_finetuned}$  & $Z_{sftmx+ae}$             \\
        \cdashline{1-5} \\
        \multicolumn{1}{l}{}             &  Anomaly Measure               &                        \multicolumn{3}{l}{CIFAR 100 (no. of known classes = 80)}                         \\
        \cdashline{1-5} \\

        A.D            & $H(y|x)$               & 0.681                   $\pm$ 0.005                   & 0.689                   $\pm$ 0.004                   & 0.682                   $\pm$ 0.007                   \\
         (Nominal Rep)     &  $Max\  P_{sftmx}(y|x)$ & 0.675                   $\pm$ 0.005                   & 0.682                   $\pm$ 0.005                   & 0.676                   $\pm$ 0.007                   \\
                           &  $Max\  P_{bce}(y|x)$   & 0.628                   $\pm$ 0.013                   & 0.685                   $\pm$ 0.006                   & 0.616                   $\pm$ 0.014                   \\
                           & $Max\ C_{\theta}(x)$                    & 0.717                   $\pm$ 0.008                   & 0.685                   $\pm$ 0.006                   & 0.724                   $\pm$ 0.009                   \\
                           & $a_{iforest}$              & 0.510                   $\pm$ 0.009                   & 0.417                   $\pm$ 0.019                   & 0.462                   $\pm$ 0.048                   \\
                           & $a_{lof}$                  & 0.441                   $\pm$ 0.014                   & 0.413                   $\pm$ 0.012                   & 0.413                   $\pm$ 0.036                   \\
        \cdashline{1-5} \\
        
        \multicolumn{1}{l}{Oracle AD}       & $g(x)$ (binary)      & 0.789                           $\pm$ 0.011                        & 0.789                           $\pm$ 0.007                        & 0.825                              $\pm$ 0.012                        \\
        (Nominal Rep)                    & $h(x)$ (multiclass)        & 0.803                           $\pm$ 0.013                        & 0.805                           $\pm$ 0.007                        & 0.828                              $\pm$ 0.014                        \\
        \cdashline{1-5} \\
        
    \multicolumn{1}{l}{Oracle AD} & $g(x)$(binary)              & 0.809    $\pm$ 0.011                   & 0.799              $\pm$ 0.011                   & 0.845                   $\pm$0.012                   \\
        (Oracle Rep)      & $h(x)$ (multi-class)                & 0.822                 $\pm$ 0.011                   & 0.812                 $\pm$0.011                   & 0.857                  $\pm$ 0.011                  \\

    \hline
    \end{tabular}
\end{table}

\begin{landscape}

\begin{table}[H]
    \centering
    \caption{Oracle Analysis of CIFAR10}
    \label{oracle-cifar10}
    \vskip 0.15in
    \begin{tabular}{cl|lllll}
    \hline
        
        \multicolumn{1}{l}{}             &Training loss  & Softmax                                                       & BCE                                                       & Hybrid
        & Triplet                                          & Triplet-1    \\
        \multicolumn{1}{l}{AD Treatment} &                 & $Z_{sftmx}$  & $Z_{bce}$  & $Z_{sftmx+ae}$ & $Z_{triplet}$          & $Z_{triplet-1}$            \\
        \cdashline{1-7} \\
        \multicolumn{1}{l}{}             &  Anomaly Measure               &                        \multicolumn{5}{l}{CIFAR 10 (no.of.known classes = 6)}                         \\
        \cdashline{1-7} \\
        A.D                          & $H(p(y|x))$       & 0.742                           $\pm$ 0.018                        & 0.765                           $\pm$ 0.008                        & 0.742                              $\pm$ 0.010                        & N/A                                             & N/A                                        \\
        (Nominal Rep)                    & $Max\  P_{sftmx}(y|x)$   & 0.734                           $\pm$ 0.017                        & 0.679                           $\pm$ 0.010                        & 0.738                              $\pm$ 0.010                        & 0.686  $\pm$ 0.016                                             & 0.702  $\pm$ 0.019                                            \\
                                
                                        & $Max\  P_{bce}(y|x)$     & 0.756                           $\pm$ 0.023                        & 0.767                           $\pm$ 0.007                        & 0.778                              $\pm$ 0.005                        & N/A                                             & N/A                                           \\
                                        &  $Max\ C_{\theta}(x)$             & 0.776                           $\pm$ 0.026                        & 0.767                           $\pm$ 0.008                        & 0.787                              $\pm$ 0.009                        & N/A                                             & N/A                                             \\
                                        & $a_{iforest}$      & 0.593                           $\pm$ 0.056                        & 0.604                           $\pm$ 0.011                        & 0.491                              $\pm$ 0.125                        & 0.679 $\pm$ 0.021                                 & 0.428 $\pm$ 0.034 \\
                                        & $a_{lof}$          & 0.443                           $\pm$ 0.016                        & 0.553                           $\pm$ 0.010                        & 0.443                              $\pm$ 0.032                        & 0.660 $\pm$ 0.014                                 & 0.462  $\pm$ 0.010 \\
                                       
        \cdashline{1-7} \\
        \multicolumn{1}{l}{Oracle AD}       & $g(x)$ (binary)      & 0.905                           $\pm$ 0.015                        & 0.895                           $\pm$ 0.014                        & 0.983                              $\pm$ 0.004                        & 0.880 $\pm$ 0.010                                 & 0.917 $\pm$ 0.011 \\
        (Nominal Rep)                    & $h(x)$ (mulitclass)        & 0.810                           $\pm$ 0.018                        & 0.786                           $\pm$ 0.011                        & 0.851                              $\pm$ 0.011                        & 0.775 $\pm$ 0.008                                 & 0.818 $\pm$ 0.015 \\
        \cdashline{1-7} \\
        
        Oracle AD                       & $g(x)$ (binary)      & 0.987                       $\pm$ 0.003                & 0.986                         $\pm$ 0.003                & 0.997                    $\pm$ 0.001                & 0.985                    $\pm$ 0.003                & 0.981                       $\pm$ 0.003                \\
        (Oracle Rep) & $h(x)$ (multi-class) & 0.907                      $\pm$ 0.010                & 0.916                          $\pm$ 0.010                & 0.933                       $\pm$ 0.008                & 0.906                      $\pm$ 0.009              & 0.897                        $\pm$0.011          \\     

    \hline
    \end{tabular}
\end{table}

\end{landscape}



\subsection{Comparison of Anomaly Measures and Nominal Representations}
We begin by comparing the anomaly measures on the nominal representations. For CIFAR100, we were able to train representations with Softmax, BCE\_finetuned, and a hybrid that combines classification and reconstruction losses. For CIFAR10, we were also able to train the Triplet loss. 

The overall pattern across both CIFAR10 and CIFAR100 is that the hybrid loss, $L_{hybrid}$, combined with the max logit anomaly measure, $Max C_\theta(x)$, gives the best overall performance. The max predicted BCE probability measure, $Max P_{bce}$, gives performance slightly worse than the max logit on CIFAR100 and statistically identical performance on CIFAR10 across all representations. Both the max predicted Softmax measure and the entropy measure are slightly worse on all configurations except CIFAR100 with the BCE loss, where they are tied with $Max C_\theta$ and $Max P_{bce}$. 

While the hybrid loss gives the best representations overall, the representation obtained from the softmax loss is also quite good based on the performance of the anomaly measures. 

The Isolation Forest and LOF anomaly measures performed very poorly, with AUROC values often worse than random guessing (AUROC=0.5).

\subsection{Oracle Analysis of Representations and Anomaly Measures}

We now consider the results of the other two AD treatments. The Oracle AD (Nominal Rep) treatment confirms that the hybrid representation contains slightly more information for anomaly detection than the other representations. Interestingly, the choice of oracle anomaly detector depends on the data set. For CIFAR100, a multiclass oracle anomaly detector works better than a binary oracle, which perhaps reflects the fact that it is operating over 80 known classes. In contrast, for CIFAR10, the binary anomaly detection oracle gives better results. But the conclusion from both detectors is the same: The hybrid representation is best. There also does not appear to be much difference between the softmax and BCE representations.

Finally, consider the oracle representations produced by the Oracle AD (Oracle Rep) treatment. These results also confirm that the hybrid loss (now with oracle training) still gives the most informative latent representation for anomaly detection. Is this somewhat surprising, because one might expect that the reconstruction loss would not be useful when labels are available for all of the classes. It suggests that even in closed-world supervised classification, the hybrid loss might improve classification performance. 

\subsection{Headroom analysis}
An advantage of the oracle methodology is that it gives us an upper bound on the amount of improvement that could be gained by better anomaly measures and by better representations. Let us focus on the hybrid representation. On CIFAR100, oracle anomaly detection improves the AUROC from 0.724 to 0.828, which is a big improvement of 0.104. Oracle representation further improves the AUROC to 0.857, which is a smaller improvement of 0.029. On CIFAR10, oracle anomaly detection improves the AUROC from 0.787 to 0.983, which is an even larger improvement of 0.196. Oracle representation further improves the AUROC to 0.997, a small improvement of 0.014. This suggests that future research should focus on improving the anomaly measures rather than the latent representations. However, the low oracle anomaly detection AUROC of 0.828 on CIFAR100 suggests that additional representation improvements are needed as the number of known classes increases.

\subsection{Regularized auto-encoders for anomaly detection}

We encountered difficulties in training autoencoders for anomaly detection, so we have not included them in the analysis above. In this section, we summarize the results were able to obtain. Tables \ref{MNIST-autoencoder} and \ref{Cifar10-autoencoder} show the results of our experiments with  autoencoders on MNIST and CIFAR10. We report results for regularized $\lambda > 0$ and unregularized $\lambda = 0$ configurations. We were only able to run these experiments for one known-unknown partition of the classes. 

Table \ref{MNIST-autoencoder} illustrates that regularization is critical in case of MNIST, as the performance improves from AUROC = 0.52  without regularization to AUROC = 0.99 with regularization. But when we apply the same methodology to CIFAR10,  Table \ref{Cifar10-autoencoder} shows that regularization gives slightly worse performance for $a_{mse}$. This raises the question when autoencoders produce meaningful latent representations. As the results were poor for CIFAR10, we decided not to focus on generative representations and reconstruction error in other experiments. 


\cite{pmlr-v80-alemi18a} demonstrate that obtaining a good latent representation that can produce good reconstructions is challenging under the current autoencoder training methodology. 
There are certain domains where we are able to obtain good latent representations and reconstructions. Examples of such domains include celebA and visual input based latent space models in single-task reinforcement learning \cite{Hafner2020Dream}. One possible reason why current training mechanisms work reasonably in celebA but fail in more complex datasets could be that the images in celebA have very simple backgrounds, whereas the images in more complex data sets, such as CIFAR10, CIFAR100, and ImageNet, the image background is often very complex.


\begin{table}[H]
\begin{center}
    \caption{MNIST autoencoder anomaly detection performance}
    \label{MNIST-autoencoder}
    \vskip 0.15in
    \begin{tabular}{l|l|l}
    \hline
    Dataset                   & \multicolumn{2}{c}{MNIST}                            \\ \hline
    Known Classes             & \multicolumn{2}{c}{0,1,2,3,4,5,8}               \\ 
    Unknown Classes           & \multicolumn{2}{c}{3,6,7,9}                   \\ 
    $\lambda$   & 0.001                       & 0.0    \\ \cdashline{1-3}
                              & \textbf{AUROC}            & \textbf{AUROC}            \\ \cdashline{1-3}
    $a_{mse}$  & 0.96 & 0.98 \\ 
    Iforest(classwise)         & 0.99 & 0.52 \\ 
    \hline
    \end{tabular}
\end{center}
\end{table}


\begin{table}[H]
    \begin{center}
        \caption{CIFAR10 autoencoder anomaly detection performance}
        \label{Cifar10-autoencoder}
        \vskip 0.15in
        \begin{tabular}{l|l|l}
        \hline
        Dataset                   & \multicolumn{2}{c}{CIFAR10}                    \\ \hline
        Known Classes             & \multicolumn{2}{c}{0,1,2,3,4,5,}               \\ 
        Unknown Classes           & \multicolumn{2}{c}{6,7,8,9}                   \\ 
        $\lambda$   & 0.00001                       & 0.0    \\ \cdashline{1-3}
                                  & \textbf{AUROC}            & \textbf{AUROC}            \\ \cdashline{1-3}
        $a_{mse}$ & 0.58 & 0.61 \\ 
        Iforest(classwise)         & 0.52 & 0.50 \\ 
        \hline
        \end{tabular}
    \end{center}
    \end{table}
    

\subsection{Classification Performance}
Although our goal is not to produce state-of-the-art classifiers, we have found that high-performing classifiers produce better representations for anomaly detection (and vice versa). Therefore, we report the classification performance of our models. On CIFAR10, the classification accuracy of the learned representations was around $86 \pm 2 \%$ accuracy. The highest accuracy was attained with the hybrid loss that combines softmax and reconstruction. It achieves an accuracy of $88\%$.

On CIFAR100, the classification accuracy of most representations was around $56\%$. Again the hybrid method achieved slightly better accuracy of $58\%$. This performance is much lower than the published state-of-the-art, which is around 75\%. It may be that additional hyper-parameter tuning could improve this number. However, the state-of-the-art is obtained using a larger data set and performing data set augmentation, so 58\% performance may be very good for our setting.

\section{Discussion}

An interesting observation from Table \ref{oracle-cifar10} is that anomaly detection based methods such as IForest and LOF achieved their best performance on $Z_{triplet}$. A possible reason for this is that both of these methods rely on distance metrics. LOF computes Euclidean distances, and IForest scores are related to the L1 distances between points \cite{10.5555/3045390.3045676}. This suggests that it may be important to match the properties of the representation to the requirements of the anomaly measure. Recent work such as \cite{sun2020conditional} and \cite{zhang2020hybrid} have demonstrated a significant boost in performance on open category detection, and their performance boost could partly be because they train a representation to maximize the effectiveness of their anomaly measure, which is based on a density estimator.

It is interesting to note that on CIFAR100, the discrepancy between the nominal and oracle representations is smaller compared to CIFAR10. A possible explanation for this is that as the number of classes increases, the anomaly detection problem becomes more difficult. Consider what happens to a query that belongs to a novel class. As the number of classes grows, it becomes more likely that images in a novel class share features with known classes. This makes it more challenging to detect that the query is novel, because the novel features may be overwhelmed by the shared features. Further analysis of oracle anomaly detectors and representations may help study this question.

\section {Conclusions} 
We summarize our conclusions as follows:

\begin{itemize}
    \item   Representations
    
    \begin{itemize}
        \item Discriminative representations are more useful than generative representations for anomaly detection.
        \item Generative representations were sufficient for good open set detection performance on MNIST but not for richer datasets such as CIFAR10.
        \item Hybrid representations that combine discriminative and reconstruction objectives give the best performance according to all of our experiments: existing anomaly detection measures, oracle anomaly detection, and oracle representation.

    \end{itemize}
    
    \item Anomaly Signals

    \begin{itemize}
        \item The \textit{maximum class logit} is found to be most effective for open set detection.
        \item From the oracle anomaly detection experiments, we can infer that nominal representations contain more information to detect unknown queries than current anomaly measures extract. There is about $10\%$ point gap in AUROC performance between oracle AD and the best anomaly measure on CIFAR100 and a $19\%$ point gap in AUROC on CIFAR10.
        
    \end{itemize}
    
    \item Oracle AD on nominal representations indicate significant room for improvement on anomaly measure, and future research in that direction would be useful.
    
    \item As the number of known classes increases, the anomaly detection problem becomes more difficult.

\end{itemize}

\section{Limitations and Future Work}

The oracle anomaly detection results for CIFAR100 indicate that with existing methods for training deep representations there is a significant loss of information that makes anomaly detection difficult. An important direction for research is to find methods, especially for problems with a large number of classes, that can learn better deep representations. 

In theory, generative representations that minimize reconstruction error should work well, because they do not depend on learning a representation that can correctly separate the unknown classes from the known classes. Instead, they rely on the \textit{inability} of the learned representation to reconstruct images from the unknown classes. However, we were not able to train good autoencoders, and we suspect that even if we could, the pixel-level loss does not capture the semantically important aspects of images. 


This report did not consider representations from self-supervised training methods such as SimCLR \cite{chen2020simple}, Deep Infomax \cite{hjelm2018learning}, and similar work, as these are not trained explicitly using class label information. Self-supervised learning representations could to be interesting to explore for anomaly detection. 
This report also did not include methods based on ensembles. Exploring efficient methods for creating large ensembles, such as HyperGAN \cite{pmlr-v97-ratzlaff19a} is an important direction for future research.. 

\indent Ideally we want any robust machine learning systems and representations to have following properties:
\begin{itemize}
    \item Good closed set predictions.
    \item Easy to detect unknown categories, or more broadly detecting and reacting in appropriate way for different kinds of novelties.
    \item Easy to generalize and adapt under distribution shifts, efficient few shot learning and active learning.
    \item Less sensitive to Adversarial perturbations.
\end{itemize}

The performance of hybrid loss functions for representation learning in our experiments suggests that adding auxiliary tasks to the basic supervised classification task may help constrain the learned representation. \cite{arjovsky2019invariant} present an intriguing method that seeks to capture invariants in representation learning. The authors propose a new objective called ``invariant risk minimization''. This is another form of constraint that could lead to better representations. 

Many people claim that deep learning succeeds because of its ability to learn representations. Our work shows that there is still much research needed to learn representations that work well for open set classification and anomaly detection.

\section*{Acknowledgements} I would like to sincerely thank my advisors Dr. Thomas G. Dietterich and Dr. Alan Fern for their guidance over the years including this work. This research was supported by grants from The Future of Life Institute, the National Science Foundation (NSF), and three DARPA Programs: Transparent Computing, Assured Autonomy, and CAML. Any opinions, findings, and conclusions or recommendations expressed in this material are those of the author(s) and do not necessarily reflect the views of the sponsors.

\bibliographystyle{apalike} 
\bibliography{references}





\end{document}